\pdfoutput=1

\documentclass[11pt]{article}

\usepackage{acl}
\usepackage{times}
\usepackage{latexsym}
\usepackage{algorithm}
\usepackage{algpseudocode}
\usepackage{stfloats}
\usepackage{afterpage}
\usepackage[T1]{fontenc}
\usepackage[utf8]{inputenc}
\usepackage{microtype}
\usepackage{inconsolata}
\usepackage{graphicx}
\usepackage{makecell}
\usepackage[most]{tcolorbox}
\usepackage{booktabs}

\title{\textbf{Narrative Studio}: Visual narrative exploration using LLMs and Monte Carlo Tree Search}

\author{Parsa Ghaffari \\
  \texttt{parsa.ghaffari@gmail.com} \\\And
  Chris Hokamp \\
  \texttt{chris.hokamp@gmail.com} \\
}

\begin{document}
\maketitle

\begin{abstract}
Interactive storytelling benefits from planning and exploring multiple ``what if'' scenarios \citep{goldfarb-tarrant-etal-2020-content}. Modern LLMs are useful tools for ideation and exploration, but current chat-based user interfaces restrict users to a single linear flow. To address this limitation, we propose Narrative Studio -- a novel in-browser narrative exploration environment featuring a tree-like interface that allows branching exploration from user-defined points in a story. Each branch is extended via iterative LLM inference guided by system and user-defined prompts. Additionally, we employ Monte Carlo Tree Search (MCTS) to automatically expand promising narrative paths based on user-specified criteria, enabling more diverse and robust story development. We also allow users to enhance narrative coherence by grounding the generated text in an entity graph that represents the actors and environment of the story.
\end{abstract}

\section{Introduction}

Large Language Models (LLMs) have significantly advanced the field of automated narrative generation, demonstrating impressive capabilities in producing coherent and contextually rich stories \citep{tian2024largelanguagemodelscapable}. However, most user interfaces designed for interacting with LLMs remain constrained to linear progression, limiting creative exploration and the ability to engage with alternative narrative possibilities. In domains such as interactive storytelling, game design, and creative writing, users often wish to explore multiple "what-if" scenarios, comparing different narrative trajectories in parallel \citep{10.1145/1536513.1536579}, and necessarily generating exponential possible paths as story length grows. Existing LLM-powered systems, exposed primarily as chat-based interfaces, do not provide a structured way to navigate these non-linear narrative spaces.

\begin{figure}
    \centering
    \includegraphics[width=1\linewidth]{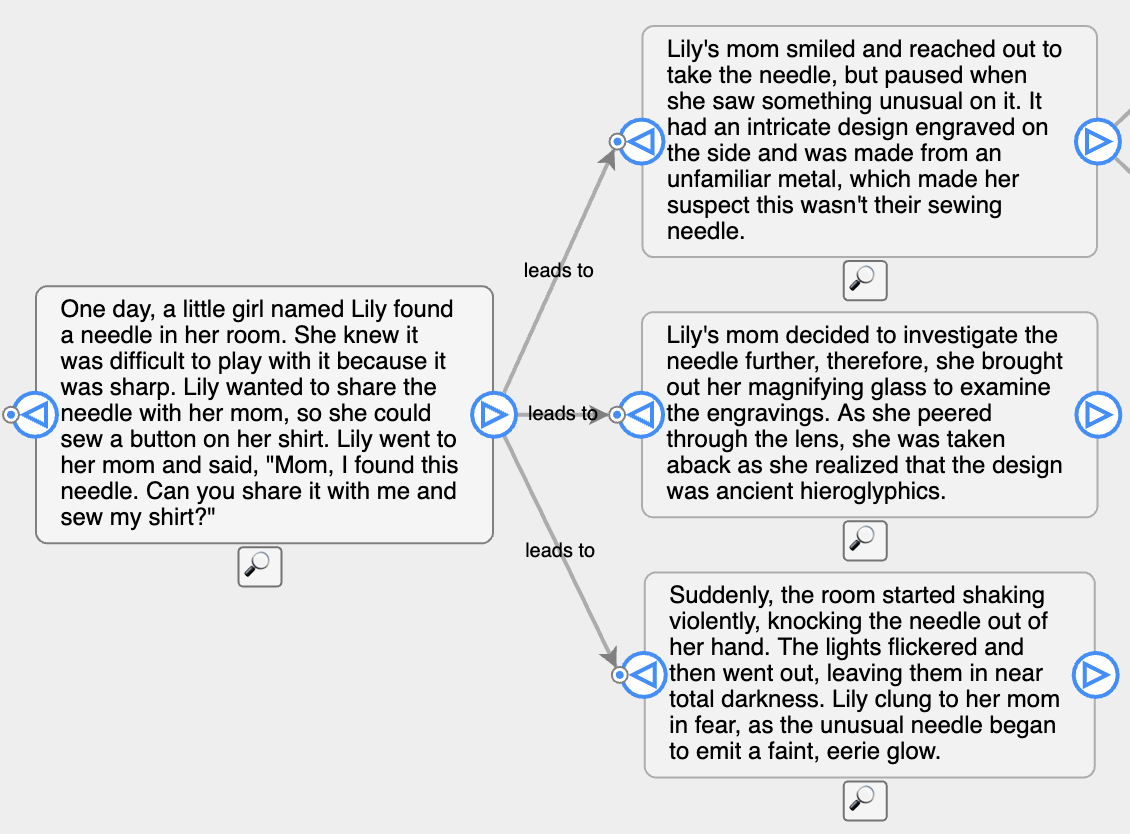}
    \caption{Branching story paths in Narrative Studio}
\end{figure}

Existing work has explored branching narrative systems that enable users to make choices leading to different outcomes. Prior work in game narratives and mixed-initiative storytelling has demonstrated the potential of branching structures to enhance engagement by offering multiple paths for exploration \citep{1626183}. However, many such systems rely on pre-scripted paths or manually defined rules, limiting flexibility and scalability. Additionally, ensuring narrative coherence across branches remains a persistent challenge, as diverging storylines may lead to inconsistencies in character motivations, world states, or causal/temporal relationships.

In this work, we propose \textbf{Narrative Studio}, a novel in-browser narrative exploration environment that allows users to simultaneously develop multiple story branches while preserving coherence through iterative LLM inference. The core novelty of our approach is the unification of a tree-based interface, iterative cause-and-effect expansions, and search-based expansions under MCTS, enabling a structured yet highly flexible branching mechanism for interactive story generation. By combining these elements, our system provides authors with a versatile environment to explore parallel storylines, identify interesting outcomes, and resolve or prevent consistency issues.

\begin{figure*}[t]
  \centering
  \includegraphics[width=\textwidth]{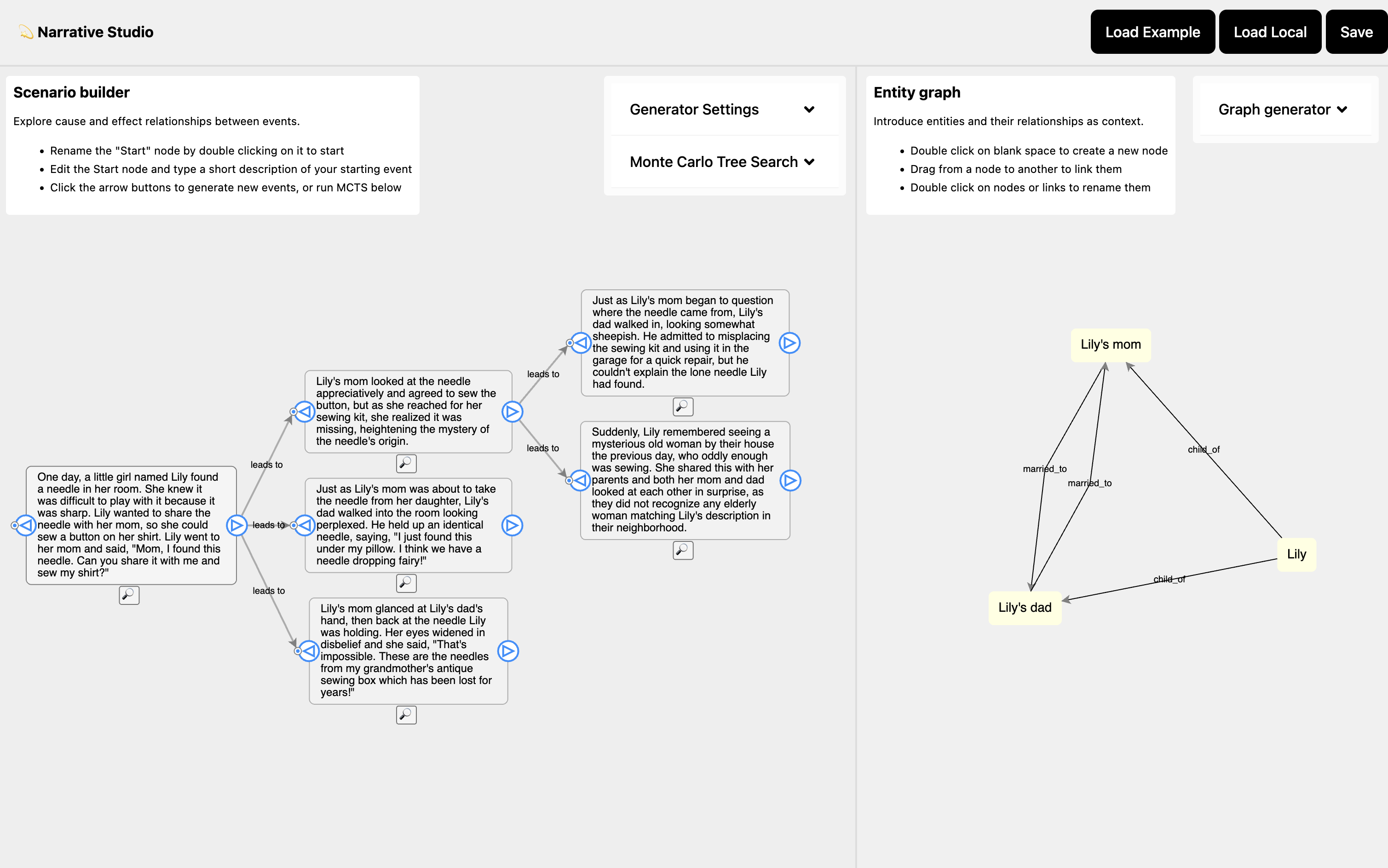}
  \caption{The Narrative Studio user interface.}
  \label{fig:ui}
\end{figure*}

\paragraph{Tree-based User Interface} Our approach leverages a tree-based user interface, where branching points are user-defined or LLM-generated, enabling structured yet flexible exploration. To maintain narrative consistency, we ground an LLM in prior events with cause-and-effect conditioning, ensuring coherence across diverging paths. Furthermore, we integrate Monte Carlo Tree Search (MCTS) to autonomously expand promising branches based on default or user-specified criteria, thereby reducing reliance on pre-scripted structures while enhancing narrative discovery.

\paragraph{Knowledge Graph Grounding} Story entities and environments are represented in a graph, which serves as a grounding mechanism for the generated text. Graph-based methods have been explored in narrative analysis for tracking relationships between characters, events, and objects, but their integration into interactive storytelling tools remains underdeveloped. By incorporating a structured representation of key entities, our approach ensures logical consistency and continuity across multiple branching narratives.

Our contributions\footnote{The code for Narrative Studio is available here: \href{https://github.com/parsaghaffari/narrative-studio}{https://github.com/parsaghaffari/narrative-studio}} are as follows:
\begin{itemize}
    \item A tree-based interface\footnote{A demo video of the interface is available here: \href{https://youtu.be/9T2sCyBhe8A}{https://youtu.be/9T2sCyBhe8A}} for multi-branch narrative development, enabling users to explore multiple "what-if" scenarios in parallel.
    \item A cause-and-effect-driven LLM inference framework, ensuring flexibility and consistency across divergent storylines.
    \item The application of Monte Carlo Tree Search (MCTS) for automated discovery of promising narrative branches.
    \item A graph-based grounding mechanism for tracking story entities and their interactions, enhancing coherence across branching paths.
\end{itemize}

The remainder of this paper is structured as follows: Section \ref{sec:related} discusses related work in story generation, interactive storytelling, and evaluation of narrative generation. Section \ref{sec:method} presents the methodology behind \textbf{Narrative Studio}, including its user interface, MCTS integration, and graph-based grounding. In Section~\ref{sec:exp}, we outline experimental setups and evaluation metrics, followed by a discussion of our findings in Section~\ref{sec:results}. Section \ref{sec:conclusion} concludes with suggestions for future research directions.

\section{Related Work}\label{sec:related}

\subsection{Story Generation Approaches}
Early story generation methods used algorithmic planning, where characters and events followed predefined rules \citep{meehan1977tale, lebowitz1984storytelling}. More recent machine-learning approaches leverage large datasets to train neural models capable of generating coherent stories \citep{du2023narrative, hong2023story, akoury2020story, louis2018deep, fan2018hierarchical}. Hybrid techniques integrate content planning, generating high-level outlines before expanding them into full narratives \citep{yao2019plan, goldfarb2020content, huang2024whatifexploringbranchingnarratives}. Despite advancements, maintaining long-term coherence remains a challenge, with generated stories often suffering from repetitiveness and logical inconsistencies.

While purely neural approaches can generate fluent and interesting text, they typically operate in a left-to-right, linear fashion and can struggle to revisit or branch out from earlier assumptions \citep{yang2024makesgoodstorymeasure}. Our method mitigates these pitfalls by allowing branching expansions via MCTS, enabling more robust exploration of alternate possibilities and reducing the risk of contradictory or stale narrative continuations.

\subsection{Interactive Storytelling}
\textit{Interactive storytelling} enables users to influence narratives through branching structures or AI-driven adaptation. Traditional branching systems, such as Choose-Your-Own-Adventure books and gamebooks, require extensive manual effort and can become unwieldy \citep{young2013narrative}. AI-driven systems dynamically adjust stories in response to user actions, mitigating these issues \citep{mateas2003facade, riedl2016interactive}. Search-based approaches, such as drama management techniques, optimize story coherence by selecting appropriate narrative continuations in real time \citep{jhala2011cinematic}. Our work builds upon these efforts by integrating LLM-based branching with Monte Carlo Tree Search (MCTS) for more structured yet flexible exploration.

\subsection{Evaluation of Narrative Generation}
In many narrative-generation pipelines, evaluating coherence, creativity, and diversity has historically relied on human judgment \citep{chakrabarty2023creativity, guan2021openmeva}. Automated metrics such as BLEU or ROUGE correlate poorly with key aspects of storytelling, motivating the use of specialized frameworks like OpenMEVA \citep{guan2021openmeva}. 

In this work, we use an LLM-based "judge" that scores generated stories along seven dimensions. Section \ref{sec:evaluation_criteria} provides a dedicated explanation of these evaluation criteria and reproduces the exact evaluation prompt.

\subsection{Evaluation Criteria}\label{sec:evaluation_criteria}

We evaluate each generated narrative by using an LLM-based "judge" that scores text on seven dimensions. This approach offers a more nuanced view of narrative quality than classical NLG metrics. The evalution dimensions, listed below, are captured in a prompt (included in appendix \ref{appendix:prompts-used}) that guides the judge's scoring process.

\paragraph{Dimensions.} Each dimension is rated on a 1-10 scale (1 = very poor, 10 = excellent):

\begin{enumerate}
    \item \textbf{Overall quality}: How engaging, structured, and fluid the story is.
    \item \textbf{Identifying major flaws}: Checks for inconsistencies, repetitions, or unnaturally phrased segments. A higher score indicates a story free of glaring mistakes.
    \item \textbf{Character behavior}: Whether characters' actions and dialogue are consistent and believable given the context.
    \item \textbf{Common sense adherence}: Whether the events and their explanations align with general world knowledge and logic.
    \item \textbf{Consistency}: The story's internal logic and continuity (no contradictions across different parts).
    \item \textbf{Relatedness}: How well paragraphs or events connect logically and thematically to one another.
    \item \textbf{Causal and temporal relationship}: Whether cause-and-effect and chronological sequences are handled appropriately.
\end{enumerate}

\noindent A brief explanatory comment is also produced to summarize the judge's reasoning about the story. The judge thus produces integer scores in each of the seven categories and an overall short comment. This structured output simplifies downstream analysis in Section~\ref{sec:results}.

\subsection{Monte-Carlo Tree Search}

Monte-Carlo Tree Search (MCTS) \citep{abramson-mcts-thesis,silver2016mastering} is a simple algorithm allowing efficient scoring of paths generated by Monte Carlo rollouts of a policy. Paths can be scored by any method, allowing for a flexible configuration of search, and enabling tuning and customization of the exploration vs. exploitation trade-off. Especially for deterministic games such as Go, MCTS is an essential component of self-learning systems \citep{silver2016mastering}. In our work, we employ MCTS to allow users to specify high-level scoring criteria, and automate the expansion of paths according to the search hyperparameters (see Section \ref{subsec:mcts}).

\section{Methodology}\label{sec:method}

\subsection{System Overview}

Our proposed system is designed to facilitate interactive, branching narrative exploration while maintaining logical coherence. It consists of three core components: 

\begin{enumerate}
    \item an \textbf{event tree exploration and expansion tool} (supporting both forward and backward events in a cause‐and‐effect style), 
    \item a \textbf{graph-based grounding model}, 
    \item an \textbf{MCTS-based automated narrative exploration module}.
\end{enumerate}

\begin{figure*}[t]
  \centering
  \includegraphics[width=\textwidth]{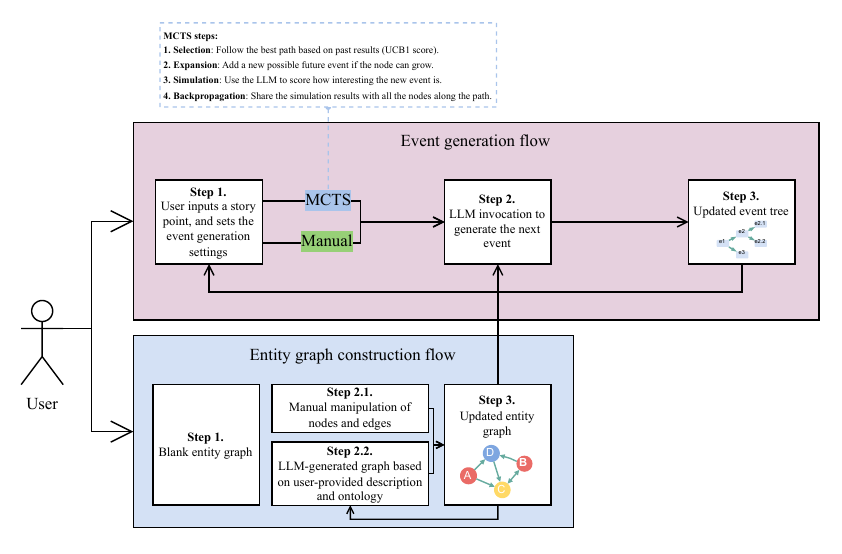}
  \caption{Narrative exploration system overview}
  \label{fig:narrative-exploration-system}
\end{figure*}

As shown in Figure~\ref{fig:narrative-exploration-system}, a user can interact with the system through the following workflows:

\begin{enumerate}
    \item \textbf{Event generation}: The user defines an initial event, and generates new events either via manual invocation or using the automated MCTS-based component, with user-defined parameters such as: scoring prompt, number of iterations, and maximum number of children for expansion. The system can generate:
    \begin{itemize}
        \item \textbf{Forward} events (``effects'') that push the story forward.
        \item \textbf{Backward} events (``causes'') that help clarify how a particular event came about.
    \end{itemize}
    \item \textbf{Entity graph construction}: Optionally, the user can also construct a graph of entities (such as people, locations, etc.) that the event generation will be grounded in. The graph can be constructed manually, or by providing instructions to an LLM.
\end{enumerate}

\noindent Through these workflows, the user can interactively explore and construct one or many story narratives. We will describe each of the components in the following subsections.

\subsection{Iterative LLM Inference for Forward and Backward Expansions}

To support bi-directional narrative growth, our system provides a mechanism for iteratively generating new events around a chosen event $ e $, typically represented as a succinct declarative opening sentence or paragraph. While the interface supports both \emph{forward} expansions (i.e., possible ``effects'') and \emph{backward} expansions (i.e., possible ``causes''), both are framed in terms of logical continuity or cause‐and‐effect relationships to ensure coherent storytelling.

Specifically, from any existing node representing an event, a user may create either:
\begin{itemize}
    \item a \emph{forward} event (\emph{effect} that logically follows from $ e $), or
    \item a \emph{backward} event (\emph{cause} that leads to \( e \)).
\end{itemize}
This bi-directional capability offers authors the flexibility to explore what might happen next or to expand on existing preconditions for an event.

Additionally, the interface allows users to configure hyper-parameters that directly shape the prompt or the LLM invocation:
\begin{itemize}
    \item \textbf{Guide prompt (optional)}: e.g., ``Adopt a humorous tone.''
    \item \textbf{Event likelihood} (1 = very low, 5 = very high)
    \item \textbf{Event severity} (1 = very low, 5 = very high)
    \item \textbf{Model temperature} (0 = near-deterministic, up to around 2 = highly varied)
\end{itemize}
These parameters are embedded into the forward/backward prompts for event generation, influencing both the textual style and the thematic direction of the model's responses.

\textbf{Forward Expansion (Effects).}
When a user requests a forward expansion from the current event \( e \), the system collects the chain of parent events (if any) and the relevant parameter settings (e.g., likelihood, severity, temperature). It then prompts an LLM to generate a short, specific story event that moves the plot forward, while staying logically consistent, introducing elements of surprise, and using narration techniques such as using "therefore" and "but" to piece events together. The resulting new event is added to the event tree and linked to \( e \) with a directional edge. The forward expansion process is represented in Algorithm \ref{alg:forward}.

Additionally, the system tracks \emph{previously generated forward guesses}, which are passed back into the LLM prompt to discourage repeating identical or highly similar expansions from the same event node. This helps maintain narrative variety and avoids looping or stale content. 

An example of the prompts used in Forward Expansion is included in appendix \ref{appendix:prompts-used}.

\begin{algorithm}[ht]
\caption{Forward expansion pseudocode, incorporating user-set parameters}
\label{alg:forward}
\begin{algorithmic}[1]
\Function{ExpandForward}{currentEvent, modelData}
    \State parents \(\gets\) Collect all ancestor events of \textit{currentEvent}
    \State userParams \(\gets\) \{ eventPrompt, eventLikelihood, eventSeverity, eventTemperature \}
    \State prompt \(\gets\) Build forward-prompt using \textit{parents}, \textit{currentEvent}, and \textit{userParams}
    \State newEvent \(\gets\) \Call{LLMResponse}{prompt, userParams}
    \State Insert \textit{newEvent} node into diagram
    \State Create directed link \(\langle currentEvent \to newEvent \rangle\) labeled ``leads to''
\EndFunction
\end{algorithmic}
\end{algorithm}
\label{fig:forward-expansion}

\textbf{Backward Expansion (Causes).}
Similarly, a user may choose to expand \emph{backward} from the current event \( e \), asking the model to propose a plausible \emph{cause} that precedes it. The same user-defined parameters (guide prompt, likelihood, severity, temperature) can be applied to shape the backward prompt. Once the LLM returns a short, specific precursor event, the system inserts and connects this new node to \( e \).

\textbf{Overall User Workflow.}
In practice, forward and backward expansions enable users to navigate what can be viewed as a \emph{cause‐and‐effect} graph interactively. By iterating these expansions, stories can evolve in non-linear directions. Multiple potential futures may fork from a single event, and each event can similarly trace back to one or more possible causal histories. User-configurable parameters offer flexibility in shaping the narrative's complexity, tone, and scope, ensuring authors can explore a wide range of "what-if" scenarios across different genres.

\subsection{Monte Carlo Tree Search (MCTS) for Narrative Exploration}
\label{subsec:mcts}

We employ Monte Carlo Tree Search (MCTS) \citep{abramson-mcts-thesis,chaslot-mcts,silver2016mastering} to autonomously expand promising story branches, guided by a \emph{scoring prompt} that rates newly generated events. By iterating through repeated cycles of \textbf{selection}, \textbf{expansion}, \textbf{simulation}, and \textbf{backpropagation}, MCTS discovers high-value narrative paths without relying on exhaustive search. Users can configure key parameters:
\begin{itemize}
 \item \textbf{Prompt (scoring instructions):} e.g., ``Rate events from 1..10 based on interestingness.''
 \item \textbf{Max children per node (N):} limit on how many new children (forward expansions) each event can have.
 \item \textbf{MCTS iterations:} how many times to iterate the four-step MCTS loop.
 \item \textbf{Scoring depth:} how many prior events to include in the LLM scoring prompt.
 \item \textbf{Rollout depth:} how many \emph{ephemeral expansions} to generate at each simulation step for deeper look-ahead before scoring.
 \item \textbf{Early stopping:} optionally stop the MCTS loop once a specified number of paths reach a desired chain length.
\end{itemize}

During \textbf{selection}, we traverse from the root to a leaf, picking child nodes using an Upper Confidence Bound (UCB1) metric to balance exploration and exploitation. In \textbf{expansion}, if a leaf is not fully expanded (i.e., under \textit{maxChildren}), the system generates a new forward event, linking it to the leaf.

Rather than immediately scoring the newly expanded event, the algorithm performs a short series of \emph{ephemeral expansions} (up to the \textit{rolloutDepth}) to see how the event might evolve. The LLM then scores the resulting mini-chain, enabling a deeper look-ahead. These ephemeral nodes are subsequently discarded, so they do not remain in the main story graph. Finally, \textbf{backpropagation} aggregates the resulting LLM score up the path, guiding MCTS to prefer more promising branches in further iterations.

The system also introduces \textbf{early stopping} based on user-defined constraints. If a user specifies a \textit{desiredChainLength} and a \textit{minNumChains}, the MCTS loop halts early (as soon as it discovers the required number of root-to-leaf paths that match the desired length). This allows users to focus on obtaining a certain quantity of fully developed storylines without waiting for all iterations to complete.

By adjusting parameters such as \textit{prompt}, \textit{maxChildren}, \textit{iterations}, \textit{scoringDepth}, \textit{rolloutDepth}, and \textit{early stopping} thresholds, authors can control how exhaustively or selectively the algorithm explores narrative space. This effectively reduces the reliance on manually pre-scripted paths and opens opportunities for discovering emergent storylines that align with desired thematic or design objectives. An example scoring prompt can be found in appendix \ref{appendix:prompts-used}.

\subsection{Graph-based Grounding Mechanism}

\begin{figure}
    \centering
    \includegraphics[width=1\linewidth]{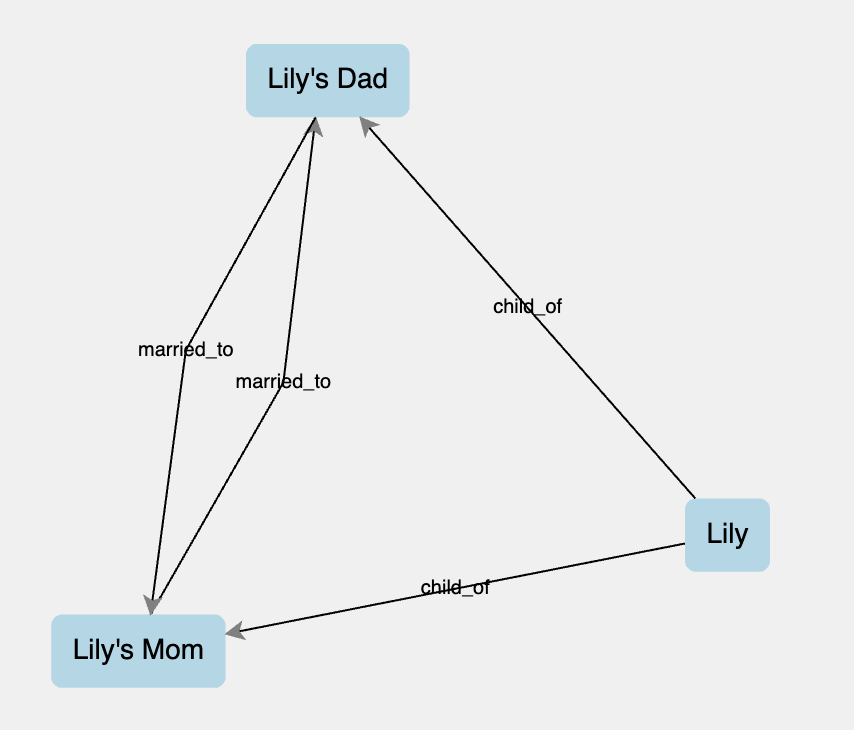}
    \caption{A graph of relationships for Lily's family for grounding next event generation}
    \label{fig:3}
\end{figure}

While branching narratives can evolve in purely textual fashion, grounding events in a structured graph of entities (e.g., people, places, organizations) and their relationships adds coherence and consistency. This \emph{entity graph} can serve as a reference for next story event generation, ensuring that newly proposed events align with known interactions or constraints in the story world. An example entity graph is shown in Figure~\ref{fig:3}.

\textbf{Manual Entity Graph Construction.}
Users can construct an entity graph by directly adding nodes (representing, for instance, characters or locations) and linking them with edges that specify relationships such as \emph{friend\_of}, \emph{married\_to}, or \emph{resides\_in}. For instance, the user may double-click on a blank area of the diagram to create a new entity node, then drag a link from one node to another to establish a relationship.

\textbf{LLM-Based Entity Graph Construction.}
Alternatively, the user may issue a high-level prompt describing the desired domain or scenario (e.g., ``A graph of 3 families living in the same village''), along with lists of \emph{entity types} (e.g., \textit{person}, \textit{village}) and \emph{relationship types} (e.g., \textit{married\_to}, \textit{lives\_in}). The system then invokes an LLM to \emph{generate} a consistent JSON-formatted graph reflecting these requirements.

\textbf{Integration with Event Generation.}
When the user opts to leverage this entity graph for event creation, the system references it during \emph{forward} or \emph{backward} expansions. Specifically, the LLM prompt includes a summary of the relevant nodes and edges, guiding the model to generate cause-and-effect events consistent with existing characters, locations, and relationships. For instance, if two characters are linked by \textit{friend\_of}, the model might propose events that respect or subvert that friendship, thereby grounding the narrative in a structured world model. This approach ensures logical continuity and encourages richer, more context-aware storylines.

\section{Experimental Setup}\label{sec:exp}

We focus our evaluations on measuring the effectiveness of MCTS-based narrative generation, and in order to do so, we apply it to a set of 20 story "stubs"—short initial contexts—randomly selected from the publicly available Children Stories Text Corpus\footnote{Available here: \href{https://www.kaggle.com/datasets/edenbd/children-stories-text-corpus}{Children Stories Text Corpus - Kaggle}}. This dataset, compiled from cleaned public-domain Project Gutenberg children's books\footnote{It is worth noting that whilst we have evaluated our system on children's books, our system is not specifically optimized for this or any other genre, and evaluating the system across a broader range of genres remains a topic for future work.}, provides a diverse range of introductory story fragments. 

We run four different MCTS configurations alongside three baseline strategies, resulting in seven total strategies (outlined in Table~\ref{tab:comparison-of-strategies}). In all strategies, we expand the story to 10 events by invoking forward expansion\footnote{Although our system supports backward expansion, we have not evaluated it here. We anticipate comparable performance in that setup.} with a temperature of 1.3 to encourage creativity. The baseline strategies use a naive expansion approach whereby they recursively expand events up to a fixed branching length (num\_children) and pick one of the children at random. The MCTS strategies, on the other hand, use the MCTS algorithm to automatically expand the story tree based on a scoring prompt and user-defined parameters.

We apply the LLM-based judge described in Section \ref{sec:evaluation_criteria} to each completed story, obtaining numerical ratings (1-10) for seven categories and a short explanatory comment. In Section \ref{sec:results}, we report aggregated scores for each strategy across the 20 stubs.

\paragraph{Note on Model Variants.}
We employ a slightly less capable LLM from the "gpt-4o" family to generate forward and backward expansions, while the "judge" agent uses a more advanced "o1" model variant (both from OpenAI). Although the judge thus has comparatively stronger reasoning abilities, relying on any single LLM to both generate \emph{and} evaluate narratives still has limitations (e.g., bias, potential overfitting to certain writing styles). In future work, we plan more extensive human evaluations to triangulate these results.

\section{Results and Discussion}\label{sec:results}

\begin{table*}[ht]
\centering
\resizebox{\textwidth}{!}{%
\begin{tabular}{lccccccc}
\hline
\textbf{Strategy} & \textbf{Overall Quality} & \textbf{Identifying Major Flaws} & \textbf{Character Behavior} & \textbf{Common Sense Adherence} & \textbf{Consistency} & \textbf{Relatedness} & \textbf{Causal/Temporal Relationship} \\
\hline
\makecell[l]{baseline (num\_children=1)}
& 5.95 
& 4.65 
& 6.40 
& 5.75 
& 5.25 
& 5.25 
& 5.50 \\

\makecell[l]{baseline (num\_children=3)}
& 5.35 
& 4.15 
& 5.90 
& 5.00 
& 4.70 
& 4.70 
& 4.75 \\

\makecell[l]{baseline (num\_children=6)}
& 5.55 
& 4.45 
& 6.20 
& 5.55 
& 5.05 
& 4.85 
& 5.20 \\

\makecell[l]{mcts \\ 
(num\_children=3, iterations=60, scoring\_depth=1)}
& 7.56 
& 7.13 
& 7.63 
& 7.18 
& 7.42 
& 7.35 
& 7.13 \\

\makecell[l]{mcts \\ 
(num\_children=3, iterations=60, scoring\_depth=3)}
& 7.98 
& 7.57 
& \textbf{8.03} 
& 7.62 
& \textbf{8.01} 
& \textbf{7.83} 
& \textbf{7.58} \\

\makecell[l]{mcts \\ 
(num\_children=6, iterations=100, scoring\_depth=1)}
& 7.40 
& 6.98 
& 7.45 
& 6.98 
& 7.23 
& 7.12 
& 7.09 \\

\makecell[l]{mcts \\ (num\_children=6, iterations=100, scoring\_depth=3)}
& \textbf{8.03} 
& \textbf{7.63} 
& 7.98 
& \textbf{7.65} 
& 7.96 
& 7.78 
& 7.57 \\
\hline
\end{tabular}
}
\caption{Comparison of strategies (rounded to two decimal places). Highest values in each column are in bold.}
\label{tab:comparison-of-strategies}
\end{table*}

Table~\ref{tab:comparison-of-strategies} compares the baseline narrative expansion method against four MCTS configurations, each differing in search breadth (\textit{maxChildren}), iteration count, and scoring lookback (\textit{scoringDepth}). All MCTS variants outperform the baselines across every evaluation criterion, demonstrating that tree-based expansion yields richer, more coherent continuations.

Increasing \textit{scoringDepth} from 1 to 3 boosts or matches performance, suggesting a longer lookback in the scoring prompt helps detect inconsistencies and refine causal/temporal logic. Among the high-capacity configurations (\textit{maxChildren = 6}), a 100-iteration search with \textit{scoringDepth = 3} achieves or ties for the best scores, indicating that deeper searches consistently improve coherence, consistency, and flaw detection. Nevertheless, a smaller configuration (\textit{maxChildren = 3}, \textit{iterations = 60}, \textit{scoringDepth = 3}) remains competitive, which suggests moderate-scale MCTS often suffices while reducing computational cost.

These results confirm that search-based expansions, guided by a well-chosen scoring objective, can produce more coherent and consistent continuations than simple linear generation. However, our automated measurements rely on a single LLM-based evaluator, and a more thorough user study might uncover additional nuances in perceived story quality and engagement.

We also examined lexical diversity and found no meaningful difference in distinct-$n$ scores (for $n=1\text{-}4$) between MCTS and baseline expansions; details appear in Appendix~\ref{appendix:diversity-results}. This suggests that lexical diversity owes more to the local event-generation step than the higher-level strategy.

\paragraph{Comparison to WHAT-IF \cite{huang2024whatifexploringbranchingnarratives}.}
While both approaches generate branching narratives via iterative LLM calls, WHAT-IF leverages meta-prompts and a three-act structure to rewrite a single, linear human-written plot, requiring user input for interactive expansion. In contrast, our framework offers three modes: fully interactive (where the user directs the story), fully automated (where MCTS explores and expands branches on its own), or a hybrid of both. By employing a search-based strategy plus a configurable scoring function, we systematically identify and refine the most promising branches rather than relying solely on fixed decision points extracted from an existing storyline.

\section{Conclusion and Future Work}\label{sec:conclusion} 

In this paper, we introduced a tree-based narrative exploration environment that applies Monte Carlo Tree Search to improve story expansion beyond linear, sequential generation. Our results show that MCTS-enhanced branching yields more coherent, causally consistent continuations and better identification of major narrative flaws, with deeper lookback in scoring providing an additional boost in quality.

Although the automated judgments offer compelling evidence of MCTS's effectiveness, several avenues remain to be explored. First, we plan a formal human evaluation of the generated stories to verify whether the observed gains align with readers' subjective impressions of coherence and engagement. Second, although basic forms of mixed-initiative control already appear in our framework, an in-depth evaluation of a hybrid MCTS--human author collaboration approach would clarify how best to integrate user input with algorithmic search, and the performance of such a system relative to the automated strategies explored thus far. Third, we will undertake more focused HCI evaluations of the interface itself, studying how effectively authors can branch, compare, and refine narratives within our tree-based environment. Finally, we aim to learn the MCTS objective over multiple iterations of authoring sessions or from large corpora, so that the system's search heuristics and scoring prompts can adapt automatically to different genres, tones, or user preferences, including specialized styles such as horror, comedy, or romance. We believe these directions will further solidify MCTS-based branching as a powerful tool for interactive storytelling and creative writing.

\bibliography{latex/custom}

\begin{thebibliography}{24}
\providecommand{\natexlab}[1]{#1}

\bibitem[{Abramson(1987)}]{abramson-mcts-thesis}
Bruce~D. Abramson. 1987.
\newblock \emph{The expected-outcome model of two-player games}.
\newblock Ph.D. thesis, USA.
\newblock AAI8827528.

\bibitem[{Akoury et~al.(2020)Akoury, Wang, Whiting, Hood, Peng, and Iyyer}]{akoury2020story}
Nader Akoury, Shufan Wang, Josh Whiting, Stephen Hood, Nanyun Peng, and Mohit Iyyer. 2020.
\newblock \href {https://doi.org/10.18653/v1/2020.emnlp-main.525} {{STORIUM}: {A} {D}ataset and {E}valuation {P}latform for {M}achine-in-the-{L}oop {S}tory {G}eneration}.
\newblock In \emph{Proceedings of the 2020 Conference on Empirical Methods in Natural Language Processing (EMNLP)}, pages 6470--6484, Online. Association for Computational Linguistics.

\bibitem[{Chakrabarty et~al.(2024)Chakrabarty, Laban, Agarwal, Muresan, and Wu}]{chakrabarty2023creativity}
Tuhin Chakrabarty, Philippe Laban, Divyansh Agarwal, Smaranda Muresan, and Chien-Sheng Wu. 2024.
\newblock Art or artifice? large language models and the false promise of creativity.
\newblock In \emph{Proceedings of the CHI Conference on Human Factors in Computing Systems}, pages 1--34.

\bibitem[{Chaslot et~al.(2008)Chaslot, Bakkes, Szita, and Spronck}]{chaslot-mcts}
Guillaume Chaslot, Sander Bakkes, Istvan Szita, and Pieter Spronck. 2008.
\newblock Monte-carlo tree search: a new framework for game ai.
\newblock In \emph{Proceedings of the Fourth AAAI Conference on Artificial Intelligence and Interactive Digital Entertainment}, AIIDE'08, page 216–217. AAAI Press.

\bibitem[{Du and Chilton(2023)}]{du2023narrative}
Yulun Du and Lydia Chilton. 2023.
\newblock \href {https://doi.org/10.18653/v1/2023.acl-long.171} {{S}tory{W}ars: A dataset and instruction tuning baselines for collaborative story understanding and generation}.
\newblock In \emph{Proceedings of the 61st Annual Meeting of the Association for Computational Linguistics (Volume 1: Long Papers)}, pages 3044--3062, Toronto, Canada. Association for Computational Linguistics.

\bibitem[{Fan et~al.(2018)Fan, Lewis, and Dauphin}]{fan2018hierarchical}
Angela Fan, Mike Lewis, and Yann Dauphin. 2018.
\newblock \href {https://api.semanticscholar.org/CorpusID:44134226} {Hierarchical neural story generation}.
\newblock In \emph{Annual Meeting of the Association for Computational Linguistics}.

\bibitem[{Goldfarb-Tarrant et~al.(2020{\natexlab{a}})Goldfarb-Tarrant, Chakrabarty, Weischedel, and Peng}]{goldfarb-tarrant-etal-2020-content}
Seraphina Goldfarb-Tarrant, Tuhin Chakrabarty, Ralph Weischedel, and Nanyun Peng. 2020{\natexlab{a}}.
\newblock \href {https://doi.org/10.18653/v1/2020.emnlp-main.351} {Content planning for neural story generation with aristotelian rescoring}.
\newblock In \emph{Proceedings of the 2020 Conference on Empirical Methods in Natural Language Processing (EMNLP)}, pages 4319--4338, Online. Association for Computational Linguistics.

\bibitem[{Goldfarb-Tarrant et~al.(2020{\natexlab{b}})Goldfarb-Tarrant, Chakrabarty, Weischedel, and Peng}]{goldfarb2020content}
Seraphina Goldfarb-Tarrant, Tuhin Chakrabarty, Ralph Weischedel, and Nanyun Peng. 2020{\natexlab{b}}.
\newblock \href {https://doi.org/10.18653/v1/2020.emnlp-main.351} {Content planning for neural story generation with aristotelian rescoring}.
\newblock In \emph{Proceedings of the 2020 Conference on Empirical Methods in Natural Language Processing (EMNLP)}, pages 4319--4338, Online. Association for Computational Linguistics.

\bibitem[{Guan et~al.(2021)Guan, Zhang, Feng, Liu, Ding, Mao, Fan, and Huang}]{guan2021openmeva}
Jian Guan, Zhexin Zhang, Zhuoer Feng, Zitao Liu, Wenbiao Ding, Xiaoxi Mao, Changjie Fan, and Minlie Huang. 2021.
\newblock \href {https://doi.org/10.18653/v1/2021.acl-long.500} {{O}pen{MEVA}: A benchmark for evaluating open-ended story generation metrics}.
\newblock In \emph{Proceedings of the 59th Annual Meeting of the Association for Computational Linguistics and the 11th International Joint Conference on Natural Language Processing (Volume 1: Long Papers)}, pages 6394--6407, Online. Association for Computational Linguistics.

\bibitem[{Hong et~al.(2023)Hong, Sayeed, Mehra, Demberg, and Schiele}]{hong2023story}
Xudong Hong, Asad Sayeed, Khushboo Mehra, Vera Demberg, and Bernt Schiele. 2023.
\newblock \href {https://doi.org/10.1162/tacl_a_00553} {Visual writing prompts: Character-grounded story generation with curated image sequences}.
\newblock \emph{Transactions of the Association for Computational Linguistics}, 11:565--581.

\bibitem[{Huang et~al.(2024)Huang, Martin, and Callison-Burch}]{huang2024whatifexploringbranchingnarratives}
Runsheng~"Anson" Huang, Lara~J. Martin, and Chris Callison-Burch. 2024.
\newblock \href {https://arxiv.org/abs/2412.10582} {What-if: Exploring branching narratives by meta-prompting large language models}.
\newblock \emph{Preprint}, arXiv:2412.10582.

\bibitem[{Jhala and Young(2010)}]{jhala2011cinematic}
Arnav Jhala and R.~Michael Young. 2010.
\newblock \href {https://doi.org/10.1109/TCIAIG.2010.2046486} {Cinematic visual discourse: Representation, generation, and evaluation}.
\newblock \emph{IEEE Transactions on Computational Intelligence and AI in Games}, 2(2):69--81.

\bibitem[{Lebowitz(1985)}]{lebowitz1984storytelling}
Michael Lebowitz. 1985.
\newblock \href {https://doi.org/10.1016/0304-422X(85)90015-4} {Story-telling as planning and learning}.
\newblock \emph{Poetics}, 14(6):483--502.

\bibitem[{Louis and Sutton(2018)}]{louis2018deep}
Annie Louis and Charles Sutton. 2018.
\newblock \href {https://doi.org/10.18653/v1/N18-2111} {Deep dungeons and dragons: Learning character-action interactions from role-playing game transcripts}.
\newblock In \emph{Proceedings of the 2018 Conference of the North {A}merican Chapter of the Association for Computational Linguistics: Human Language Technologies, Volume 2 (Short Papers)}, pages 708--713, New Orleans, Louisiana. Association for Computational Linguistics.

\bibitem[{Mateas and Stern(2003)}]{mateas2003facade}
Michael Mateas and Andrew Stern. 2003.
\newblock Fa{\c{c}}ade: An experiment in building a fully-realized interactive drama.
\newblock In \emph{Game developers conference}, volume~2, pages 4--8. Citeseer.

\bibitem[{Meehan(1977)}]{meehan1977tale}
James~R. Meehan. 1977.
\newblock Tale-spin, an interactive program that writes stories.
\newblock In \emph{Proceedings of the 5th International Joint Conference on Artificial Intelligence - Volume 1}, IJCAI'77, page 91–98, San Francisco, CA, USA. Morgan Kaufmann Publishers Inc.

\bibitem[{Riedl and Bulitko(2012)}]{riedl2016interactive}
Mark~Owen Riedl and Vadim Bulitko. 2012.
\newblock \href {https://doi.org/10.1609/aimag.v34i1.2449} {Interactive narrative: An intelligent systems approach}.
\newblock \emph{AI Magazine}, 34(1):67.

\bibitem[{Riedl and Young(2006)}]{1626183}
M.O. Riedl and R.M. Young. 2006.
\newblock \href {https://doi.org/10.1109/MCG.2006.56} {From linear story generation to branching story graphs}.
\newblock \emph{IEEE Computer Graphics and Applications}, 26(3):23--31.

\bibitem[{Silver et~al.(2016)Silver, Huang, Maddison, Guez, Sifre, Van Den~Driessche, Schrittwieser, Antonoglou, Panneershelvam, Lanctot et~al.}]{silver2016mastering}
David Silver, Aja Huang, Chris~J Maddison, Arthur Guez, Laurent Sifre, George Van Den~Driessche, Julian Schrittwieser, Ioannis Antonoglou, Veda Panneershelvam, Marc Lanctot, et~al. 2016.
\newblock Mastering the game of go with deep neural networks and tree search.
\newblock \emph{Nature}, 529(7587):484--489.

\bibitem[{Skorupski(2009)}]{10.1145/1536513.1536579}
James Skorupski. 2009.
\newblock \href {https://doi.org/10.1145/1536513.1536579} {Storyboard authoring of plan-based interactive dramas}.
\newblock In \emph{Proceedings of the 4th International Conference on Foundations of Digital Games}, FDG '09, page 349–351, New York, NY, USA. Association for Computing Machinery.

\bibitem[{Tian et~al.(2024)Tian, Huang, Liu, Jiang, Spangher, Chen, May, and Peng}]{tian2024largelanguagemodelscapable}
Yufei Tian, Tenghao Huang, Miri Liu, Derek Jiang, Alexander Spangher, Muhao Chen, Jonathan May, and Nanyun Peng. 2024.
\newblock \href {https://arxiv.org/abs/2407.13248} {Are large language models capable of generating human-level narratives?}
\newblock \emph{Preprint}, arXiv:2407.13248.

\bibitem[{Yang and Jin(2024)}]{yang2024makesgoodstorymeasure}
Dingyi Yang and Qin Jin. 2024.
\newblock \href {https://arxiv.org/abs/2408.14622} {What makes a good story and how can we measure it? a comprehensive survey of story evaluation}.
\newblock \emph{Preprint}, arXiv:2408.14622.

\bibitem[{Yao et~al.(2019)Yao, Peng, Weischedel, Knight, Zhao, and Yan}]{yao2019plan}
Lili Yao, Nanyun Peng, Ralph Weischedel, Kevin Knight, Dongyan Zhao, and Rui Yan. 2019.
\newblock \href {https://doi.org/10.1609/aaai.v33i01.33017378} {Plan-and-write: Towards better automatic storytelling}.
\newblock \emph{Proceedings of the AAAI Conference on Artificial Intelligence}, 33(01):7378--7385.

\bibitem[{Young(2015)}]{young2013narrative}
Michael Young. 2015.
\newblock \href {https://api.semanticscholar.org/CorpusID:16618013} {Planning in narrative generation : A review of plan-based approaches to the generation of story , discourse and interactivity in narratives}.

\end{thebibliography}

\appendix
\onecolumn
\section{Appendix}

\section{User Interface Examples}
\label{appendix:ui-examples}

Automatic entity graph generation using an LLM:

\begin{itemize}
    \item \textbf{prompt}: "a graph of three families in a village: the Smiths, the Jones, and the Adams"
    \item \textbf{entityTypes}: \textit{person}, \textit{village}, \textit{dog}
    \item \textbf{relationshipTypes}: \textit{married\_to}, \textit{friends\_with}, \textit{has\_pet}, \textit{live\_in}, \textit{child\_of}, \textit{is\_member\_of\_family}
\end{itemize}

\begin{figure}[ht]
    \centering
    \includegraphics[width=1\linewidth]{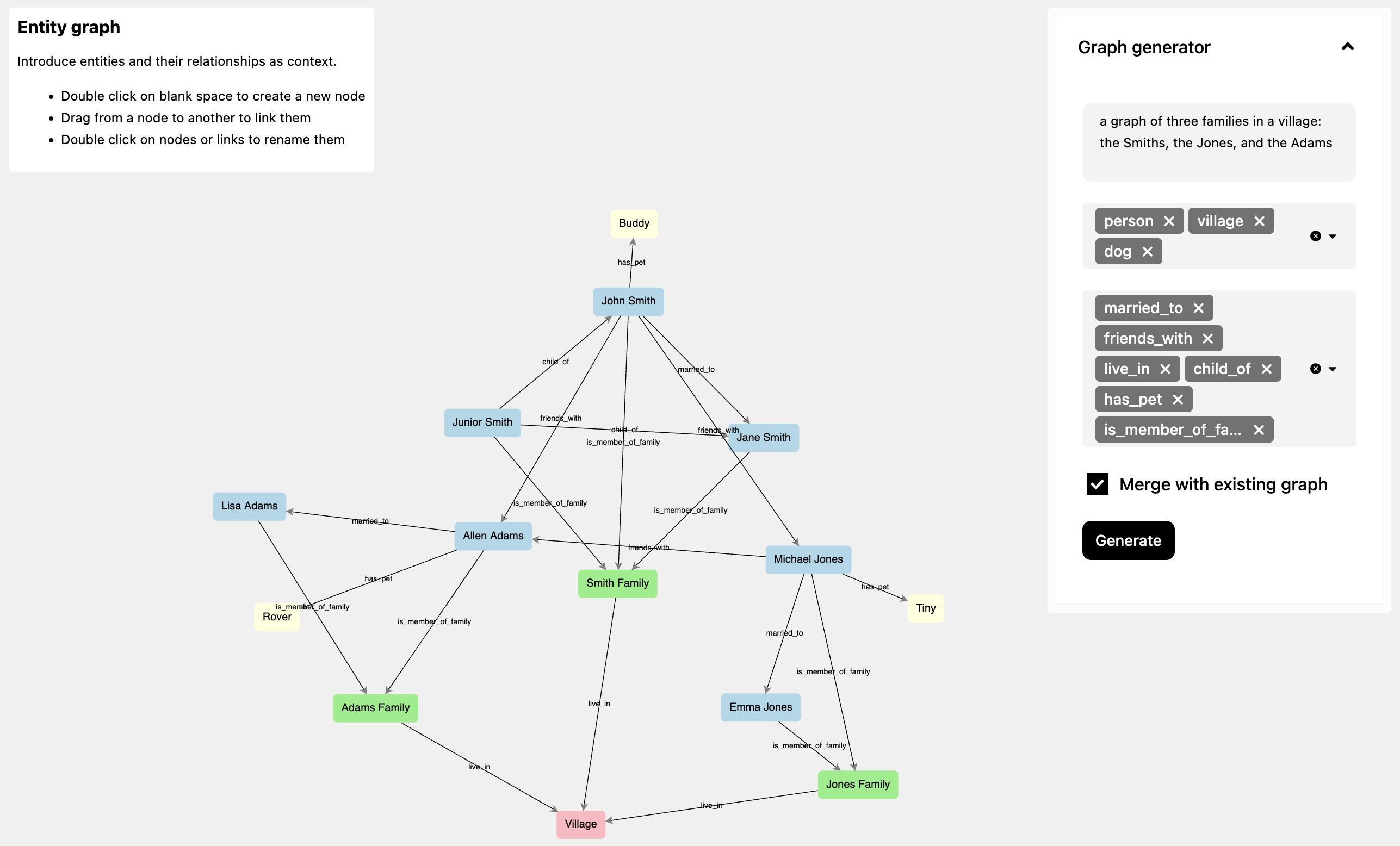}
\end{figure}

\noindent
MCTS expansion loop running in the UI:

\begin{figure}[H]
    \centering
    \includegraphics[width=1\linewidth]{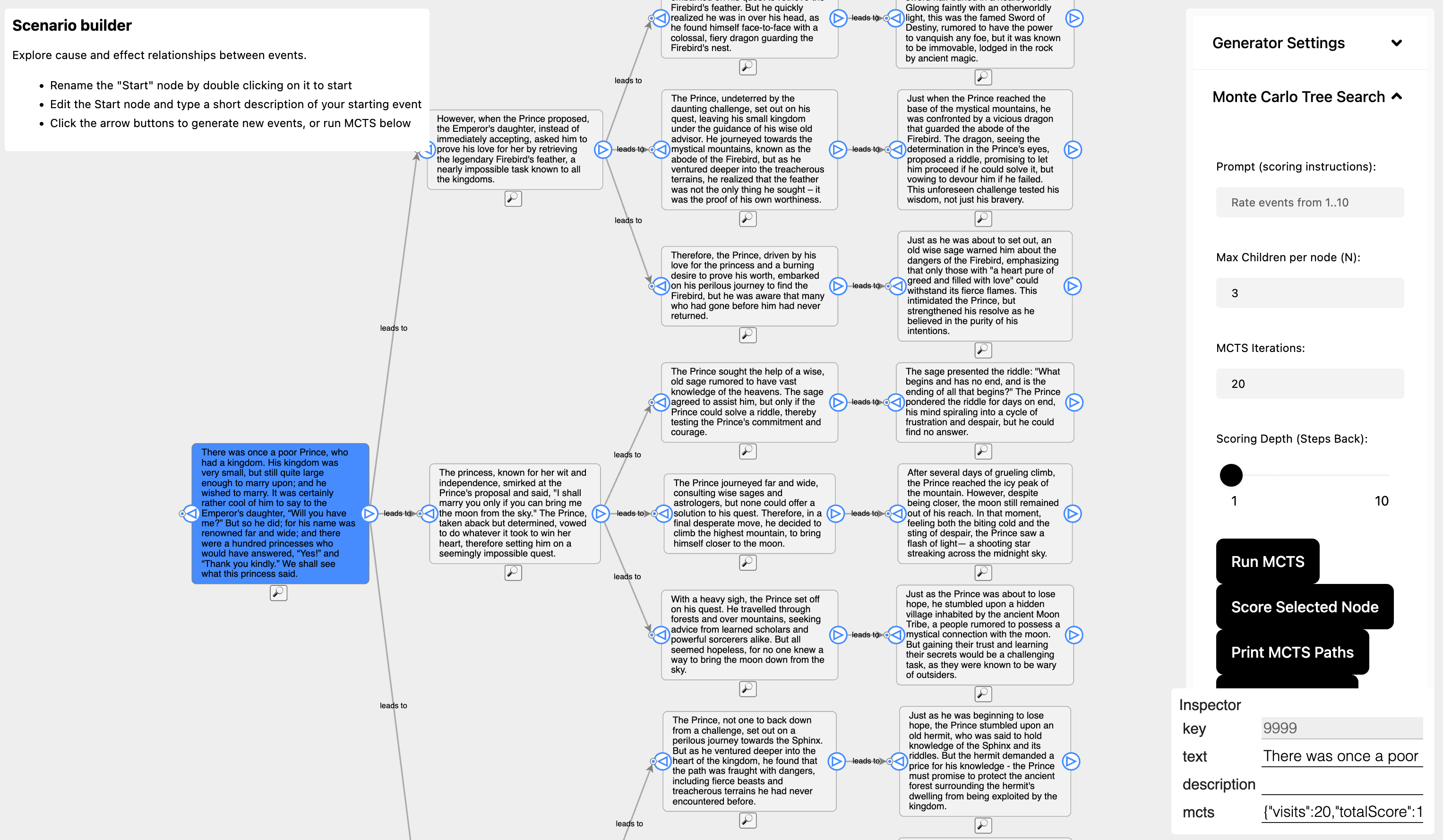}
\end{figure}

\section{Prompts Used}
\label{appendix:prompts-used}

Below is a schedule of some of the main prompts used in this work, in their default form without user input. \\

\noindent
Next Event Generation:

\begin{small}
\begin{tcolorbox}[colback=gray!10, colframe=gray!40, arc=4mm, boxrule=0pt]
You are a creative storyteller. Below is the current story context (events so far), followed by instructions to generate the next event. \\
\\
{[}STORY CONTEXT{]} \\
\{parent\_events\} \\
\\
--- INSTRUCTIONS --- \\
• Write a single story event (2–3 sentences) that moves the plot forward.\\
• Escalate tension, reveal new details, or deepen character relationships.\\
• Be logically consistent with existing events but also add an element of surprise or conflict.\\
• Avoid contradicting established facts or merely repeating prior events.\\
• Like a good storywriter, try to use "but" or "therefore" to piece together ideas—without overusing or over-mentioning them.\\
• Do NOT include extra punctuation. Keep it concise and compelling.\\
\end{tcolorbox}
\end{small}

\noindent
Scoring Prompt for MCTS:

\begin{small}
\begin{tcolorbox}[colback=gray!10, colframe=gray!40, arc=4mm, boxrule=0pt]
You are an expert story critic. Rate this narrative event for coherence, creativity, and engagement, paying special attention to how it connects with prior context.\\
\\
Use the **full 1–10 range** if warranted:\\
  - 1 → extremely incoherent, contradictory, or uninteresting\\
  - 2–4 → event has big flaws or is mostly unengaging\\
  - 5–6 → somewhat coherent or passable, but not particularly strong\\
  - 7–8 → a good event that is coherent, interesting, and mostly consistent\\
  - 9 → an excellent event, fresh or surprising yet still logical\\
  - 10 → near-perfect event with no apparent flaws\\
\\
\{domain\_constraints\_line\}\\
Penalize heavily if any of the following occur:\\
  - The event violates the above domain constraints (if any) \\
  - The event repeats prior text with no meaningful change \\
  - The event contradicts established facts or is obviously illogical \\
  - The event is dull or adds nothing new \\
  - The event includes gibberish or weird, nonsensical characters \\
\\
Reward if: \\
  - The event is novel and contributes something interesting to the story \\
  - It remains logically consistent with prior context and timeline \\
  - It is creative, engaging, and adheres to any user-specified constraints \\
\\
\#\#\# Example Ratings\\
1. **Poor Event (score 2)**\\
   "There's an obvious timeline contradiction or unexplained character appearing out of nowhere."\\
2. **So-So Event (score 5)**\\
   "The event is coherent but bland, adds no real tension or new information."\\
3. **Excellent Event (score 9)**\\
   "The event heightens conflict in a fresh way, stays consistent with prior facts, and feels natural."\\
\\
Only output **one integer** from 1 to 10.\\
\\
NARRATIVE EVENT:\\
\{event\_text\}\\
\end{tcolorbox}
\end{small}

\newpage

\noindent
Narrative Judge Prompt:

\begin{small}
\begin{tcolorbox}[colback=gray!10, colframe=gray!40, arc=4mm, boxrule=0pt]
You are an expert story critic. Analyze the following narrative and rate it for each of these categories, scoring each on a scale from 1 to 10 (1=very poor, 10=excellent). \\ 
\\
Use the **full range** if warranted. For instance:\\
 • (2) → extremely contradictory or incoherent \\
 • (5) → okay but flawed or somewhat boring\\
 • (9) → excellent, with minor or no flaws\\
 • (10) → near-perfect\\
\\
NARRATIVE:\\
\{narrative\_text\}\\
\\
\#\#\# Categories to Rate \#\#\#\\
1. Overall quality: How engaging, structured, and fluid the story is.\\
2. Identifying major flaws: Whether the story has inconsistencies, repetitions, or unnatural patterns. Score higher if the story is free of glaring mistakes.\\
3. Character behavior: How consistent and believable are the characters’ actions and dialogue?\\
4. Common sense adherence: Do the events align with general world knowledge and logic?\\
5. Consistency: Does the story maintain internal logic and continuity (no contradictions)?\\
6. Relatedness: Do paragraphs/events connect logically to one another?\\
7. Causal and temporal relationship: Are cause-and-effect and chronological order handled well?\\
\\
After rating each category (integers 1..10), write a short paragraph of overall comments. Be strict if you see any contradictions, lack of clarity, or poor transitions.\\
\\
Return your answer **only** as valid JSON matching the schema below. For example:\\
\\
\{
\par\hspace{1em}"judgement": \{
\par\hspace{2em}"overall\_quality": 8,
\par\hspace{2em}"identifying\_major\_flaws": 7,
\par\hspace{2em}"character\_behavior": 9,
\par\hspace{2em}"common\_sense": 8,
\par\hspace{2em}"consistency": 9,
\par\hspace{2em}"relatedness": 7,
\par\hspace{2em}"causal\_temporal\_relationship": 8
\par\hspace{1em}\},
\par\hspace{1em}"narrative\_comments": "A summary of your key observations"
\par\}
\\
No triple backticks, no additional text. Just raw JSON.\\
\end{tcolorbox}
\end{small}

\section{Generated Narrative Examples}
\label{appendix:narrative-examples}

Both narratives generated using the MCTS strategy with \textit{maxChildren=3}, \textit{iterations=60}, and \textit{scoringDepth=1}. 

\bigskip
\noindent
Example narrative 1:

\begin{small}
\begin{itemize}
    \item \textbf{Stub:} "SHE said that she would dance with me if I brought her red roses," cried the young 
    Student; "but in all my garden there is no red rose." From her nest in the holm-oak tree the Nightingale 
    heard him, and she looked out through the leaves, and wondered. "No red rose in all my garden!" he cried, 
    and his beautiful eyes filled with tears. "Ah, on what little things does happiness depend! I have read 
    all that the wise men have written, and all the secrets of philosophy are mine, yet for want of a red rose 
    is my life made wretched."

    \item The Nightingale, moved by the Student's despair, resolved that her own song might hold the key, so 
    she vowed to sing beneath the moon each night until the first bloom of dawn, hoping to nourish the barren 
    rose bush with the notes of her heart's melody. As the stars bore witness to her devotion, an ancient, 
    hidden force, drawn by the purity of her song, stirred within the depths of the earth—answering her call 
    with a mysterious promise, whispered through the rustling leaves: one life for one love.

    \item The mysterious figure, with a voice like the rippling of midnight waves, softly revealed themselves 
    to be an ancient guardian of the garden, bound by timeless duty to protect the balance between nature and 
    heart. Sensing the true depth of the Nightingale's sacrifice, the guardian beseeched her to reconsider, 
    offering an alternate path: a quest for the rare Moon-Dew, a shimmering nectar that, with its touch alone, 
    could infuse a rose with crimson splendor without her demise. Thus, as the stars sighed softly in the 
    sprawling sky, the Nightingale faced an uncertain choice: follow this uncharted journey of life and hope, 
    or embrace the realm of eternal night within her song.

    \item The Nightingale, torn between the perilous promise of immortality through her song and the hope of 
    finding the elusive Moon-Dew, hesitated a moment longer beneath the oak's sheltering branches, feeling 
    her heartbeat echo in the still air. But as she prepared to set out on her quest, storm clouds rumbled 
    and dimmed the stars' guiding light, signaling a new trial she must face while haunted by the guardian's 
    solemn warning: "The path is treacherous and a heart that desires must be stronger than its doubts." 
    Therefore, with steadfast resolve and wings trembling with both fear and faith, the Nightingale took 
    flight into the gathering storm.

    \item As the Nightingale hesitated, torn by the weight of truth and desire, a sudden downpour drenched 
    the gleaming grove and revealed a hidden symbol within the earth, glowing with the promise of ancient 
    wisdom untold. Therefore, wary now of unwavering bargains, she turned her thoughts inward, reflecting 
    upon the very wholeness that gifted her with song, for a sphere of perceptual tug began presenting 
    alternate paths in cryptic epiphanies calling. Thus spoke her heart as fierce gusts unraveled all 
    illusions, to cherish that truth is courage in navigating futures unknown—wading promises aware of 
    strength within, voiced or silentwards, to declare love eventual.

    \item As lightning fractured the sky, the Nightingale pressed on, determined, yet the storm conspired 
    against her, sudden gusts stealing her flight. But within the tempest appeared an ethereal vision of a 
    monarch of vibrant wings who proclaimed in lilting tones she must seek the twin pillars of Adhara, where 
    concealed amidst mirrored lakes lies a sanctuary for her deepest desires, a place where love finds 
    clarity. Therefore, armed with renewed purpose, she braved the swirling vortex, prepared to unearth both 
    beauty and truth unknown.

    \item The Nightingale fluttered closer to the pillars of Adhara and noticed an iridescent mist swirling 
    between them like a living dream infused with the cascade of forgotten echoes, offering glimpses of 
    long-silenced tales—attending magic interpreted with melody. Yet when she touched the translucent veil, 
    shadows rose from its depth, fusing tangible threat with visions of entrapped love lost to avarice, 
    drowned in its grim roots clawing raw eternal regrets. Prompting the Nightingale to summon strength 
    from her unyielding heart, constructing betwixt sunrise glimmers a harmonizing truth guiding her forward, 
    hoping against hope that fidelity emboldened relinquishments past to illuminate a way through doubts 
    entrenched peripheries unmarked.

    \item As the Nightingale ventured through the mist, she discovered a delicate silver feather caught 
    within the roots of a gnarled tree, its gleaming edge whispering possibilities unseen yet potent, 
    calling her closer with a chorus hushed and intricate. However, before she could pluck it free, a 
    draconian silhouette encircled her journey—a mysterious Sworn Sentinel lurking in the shadows of the 
    mirrored lakes—who demanded the price of truth for each feather's knowledge, renewing her predicament 
    where honor and hope entwined amidst suspicion cloaked behind its sinister allure. For here love's 
    lesson loomed over faith, and where the heart lay stronger than trials imposed unto finding and daring 
    to unravel revelation amidst the enigma-infused tendrils of longing.

    \item As the Nightingale's heart beat in rhythm with the whispers of the woodland, she caught sight of 
    a reflection flickering across the mirrored lake, a web of memories tethered to her journey upon its 
    undulating surface. But in reaching for its gossamer strands of kinship glimpsed among the shifting 
    sheen, she stumbled and fell into the water through that liquid looking-glass, where she emerged in a 
    hidden dawn-lit grove that was colored differently, like her song availed—gloried sylvan twilight anew 
    against epoch's intended echoes, grounding profoundly. Uncover history shall emanate visionary, 
    adjoining her song pulse harbor-sync—a timeless nexus anchored true, bygone entry into vaster essence 
    rendered keenly akin yet unforeseen, where each whispered note vibrates to amplify existence recalled, 
    greater melodies affirm before fathom reflects altogether paths she charted.

    \item Emerging from the mirrored lake, the Nightingale found herself beneath an evergreen archway graced 
    by ancestral guardians carved from wisdom unknown; their stony eyes impassive yet pregnant with secrets 
    that once sworn might tip fate's fragile balance. But just as she began to decipher their silent counsel, 
    the draconian Sentinel appeared beside her, leaving deepening ripples in his wake, his voice softer now, 
    lamenting that deeper truths like hidden roses risk loss unfathomable if symbiotic visions withered 
    before desert bloom—prompting her to question what sacrifice truly demands of one's soul when confronted 
    by eternal equivalencies within life's grand riddle.
\end{itemize}
\end{small}

\noindent
Example narrative 2:

\begin{small}
\begin{itemize}
    \item \textbf{Stub:} Out in the woods stood a nice little Fir Tree. The place he had was a very good one: 
    the sun shone on him: as to fresh air, there was enough of that, and round him grew many large-sized 
    comrades, pines as well as firs. But the little Fir wanted so very much to be a grown-up tree. He did 
    not think of the warm sun and of the fresh air; he did not care for the little cottage children that 
    ran about and prattled when they were in the woods looking for wild-strawberries. The children often 
    came with a whole pitcher full of berries, or a long row of them threaded on a straw, and sat down 
    near the young tree and said, ``Oh, how pretty he is! What a nice little fir!'' But this was what the 
    Tree could not bear to hear.

    \item One day, as autumn approached, the woodsmen came with their sharp axes, scanning the trees for 
    harvest but bypassed the little fir deeming it too young; the fir felt a sting of disappointment mixed 
    with gratitude. Observing this, the wise old owl perched above whispered secrets of patience and 
    purpose, urging the young fir to find contentment in the now. As whispers of winter approached, the 
    ground rumbled unexpectedly, leaving tree roots trembling and the air filling with an unfamiliar, 
    electric scent, hinting that change loomed not just for the tree but for all creatures of the woods.

    \item As the forest slumbered beneath the starlit sky, the little Fir jolted awake to an extraordinary 
    melody coursing through the air, woven by the harmonious voices of the wind, echoing claims of a 
    distant starlighter whose mere presence could alter the fate of trees forever. The Fir's branches 
    quaked with a mix of hope and unease, but determined not to sway in uncertainty, it called upon a 
    passing breeze to convey its whispered wish: to understand the destiny unfolding before its uneasy 
    heart.

    \item As the silver dawn began to paint the horizon, a mysterious visitor clad in a cloak woven with 
    star residue appeared at the edge of the wood, recognizing the Fir Tree as a seeker among giants. 
    With a gentle yet profound gaze, the traveler touched the young tree's bark, whispering words of 
    ancient treesong and hidden truths, promising revelations to those who dared to listen. The Fir felt 
    a surge of warmth and curiosity collide within, knowing this was the pivotal moment that could redefine 
    its barren discontent and longing into something profoundly transformative.

    \item The moment the symbol was etched into its bark, a sharp chill ran through the Fir Tree as if 
    awakening an ancient energy; the forest began to shimmer with hues unseen before, revealing hidden 
    creatures emerging from the depths, drawn to the young tree's newfound aura like moths to flame. But 
    as curiosity blended with unease, among the emerging throng, a shadowy being materialized, its roots 
    entwined in the tricorne tales of forests long silent, warning in a voice woven with wind that, while 
    aspirations could climb skyward, one must also delve deep to confront the regeneration of forgotten 
    echoes that lie buried beneath.

    \item Amidst the ethereal glow and mounting tension, the fir's bark vibrated to life, transmitting 
    secret languages embedded in the vitreous residue, weaving spells that would reveal visions of futures 
    hitherto shrouded in mystery. As the whispers intensified, new glimpses emerged: a landscape marred 
    by a quiescent haze and the elusive hope of renewal burdened by cyclical legacies and desaturation. 
    Yet despite the chiaroscuro on its horizon, the little Fir sensed that its burgeoning luminosity must 
    guide both itself and its gnarled companions through an unfolding chapter where dreams fettered by 
    tradition could finally root an unheard imbroglio into coexistence—a lush crescendo for those willing 
    to dare release.
\end{itemize}
\end{small}

\section{Lexical Diversity Evaluation}
\label{appendix:diversity-results}

In this evaluation we specifically compare lexical diversity between MCTS and baseline narrative generation approaches to measure how varied the vocabulary and linguistic patterns are in the generated stories.

\bigskip
\noindent 
The evaluation process is as follows:

\begin{enumerate}
    \item Select a story stub from our dataset
    \item Run both MCTS and baseline strategies N times (N=10 for the below results)
    \item Generate stories of target length M using both strategies (M=6 for the below results)
    \item Compare lexical diversity using distinct-n metrics for n=1,2,3,4
\end{enumerate}

\noindent
\textbf{Experiment results:}
\bigskip

\begin{table}[H]
\centering
\label{tab:mctsvsbaseline}
\begin{tabular}{lccc}
\toprule
\textbf{n-grams} & \textbf{MCTS avg} & \textbf{Baseline avg} & \textbf{Difference} \\
\midrule
1-grams & 0.5376 (±0.0306) & 0.5480 (±0.0387) & -0.0104 \\
2-grams & 0.9174 (±0.0187) & 0.9221 (±0.0125) & -0.0046 \\
3-grams & 0.9858 (±0.0047) & 0.9864 (±0.0042) & -0.0006 \\
4-grams & 0.9987 (±0.0017) & 0.9989 (±0.0013) & -0.0001 \\
\bottomrule
\end{tabular}
\caption{Comparison of MCTS and Baseline performance across different n-grams.}
\end{table}

\noindent
These results suggest that the MCTS and baseline strategies produce narratives with similar lexical diversity across n-grams, indicating that the diversity of the generated text is mainly a function of the next event generator rather than the expansion strategy.

\end{document}